\documentclass[10pt,twocolumn,letterpaper]{article}
\pdfoutput=1

\usepackage{iccv}
\usepackage{times}
\usepackage{epsfig}
\usepackage{graphicx}
\usepackage{amsmath}
\usepackage{amssymb}
\usepackage{color}

\usepackage{tabulary}
\usepackage{tabularx}
\usepackage{booktabs}
\usepackage{stfloats}
\usepackage{multicol}
\usepackage[ruled,vlined]{algorithm2e}

\newcommand{\ignore}[1]{}

\makeatletter
\DeclareRobustCommand\onedot{\futurelet\@let@token\@onedot}
\def\@onedot{\ifx\@let@token.\else.\null\fi\xspace}

\makeatother

\definecolor{MyDarkBlue}{rgb}{0,0.08,1}
\definecolor{MyDarkGreen}{rgb}{0.02,0.6,0.02}
\definecolor{MyDarkRed}{rgb}{0.8,0.02,0.02}
\definecolor{MyDarkOrange}{rgb}{0.40,0.2,0.02}
\definecolor{MyPurple}{RGB}{111,0,255}
\definecolor{MyRed}{rgb}{1.0,0.0,0.0}
\definecolor{MyGold}{rgb}{0.75,0.6,0.12}
\definecolor{MyDarkgray}{rgb}{0.66, 0.66, 0.66}

\newcommand{\norm}[1]{\left\lVert#1\right\rVert}

\newcommand{\myparagraph}[1]{\vspace{-2pt}\paragraph{#1}}

\newcommand{\ra}[1]{\renewcommand{\arraystretch}{#1}}
\usepackage{multirow}

\usepackage[pagebackref=true,breaklinks=true,letterpaper=true,colorlinks,bookmarks=false]{hyperref}

\iccvfinalcopy 

\def\iccvPaperID{3505} 

\ificcvfinal\fi

\begin{document}

\title{\vspace{-3em}SurfGen: Adversarial 3D Shape Synthesis with Explicit Surface Discriminators}

\author{Andrew Luo\textsuperscript{1} \quad Tianqin Li\textsuperscript{1} \quad Wen-Hao Zhang\textsuperscript{2} \quad Tai Sing Lee\textsuperscript{1} \\ 
\textsuperscript{1}Carnegie Mellon University\qquad \textsuperscript{2}University of Chicago
}

\maketitle
\ificcvfinal\thispagestyle{empty}\fi


\begin{abstract}
    Recent advances in deep generative models have led to immense progress in 3D shape synthesis. While existing models are able to synthesize shapes represented as voxels, point-clouds, or implicit functions, these methods only indirectly enforce the plausibility of the final 3D shape surface. Here we present a 3D shape synthesis framework (SurfGen) that directly applies adversarial training to the object surface. Our approach uses a differentiable spherical projection layer to capture and represent the explicit zero isosurface of an implicit 3D generator as functions defined on the unit sphere. By processing the spherical representation of 3D object surfaces with a spherical CNN in an adversarial setting, our generator can better learn the statistics of natural shape surfaces. We evaluate our model on large-scale shape datasets, and demonstrate that the end-to-end trained model is capable of generating high fidelity 3D shapes with diverse topology. Code is available at \url{https://github.com/aluo-x/NeuralRaycaster}. 
\end{abstract}
\section{Introduction}

Shape generation is primarily concerned with the synthesis of diverse, realistic, and novel shapes. High fidelity models of 3D shapes are key to creating immersive virtual worlds, and are important to many disciplines, including architecture, visual effects, and training robots in simulated navigation. 


With the introduction of large scale 3D object datasets such as ShapeNet~\cite{chang2015shapenet} and ModelNet~\cite{wu20153d}, there has been significant progress towards building generative models of 3D shapes. The predominant approach is to perform training using representations that are straightforward for neural networks to work with, such as voxel grids, point clouds, or implicit functions. This is in contrast to the majority of graphics \& simulation tasks, which require an triangle mesh representation of a shape. When learned in an adversarial setting, the discriminator in these frameworks only indirectly ensure the realism of the object surface.

\begin{figure}[t!]
  \centering
  \includegraphics[width=0.9\linewidth]{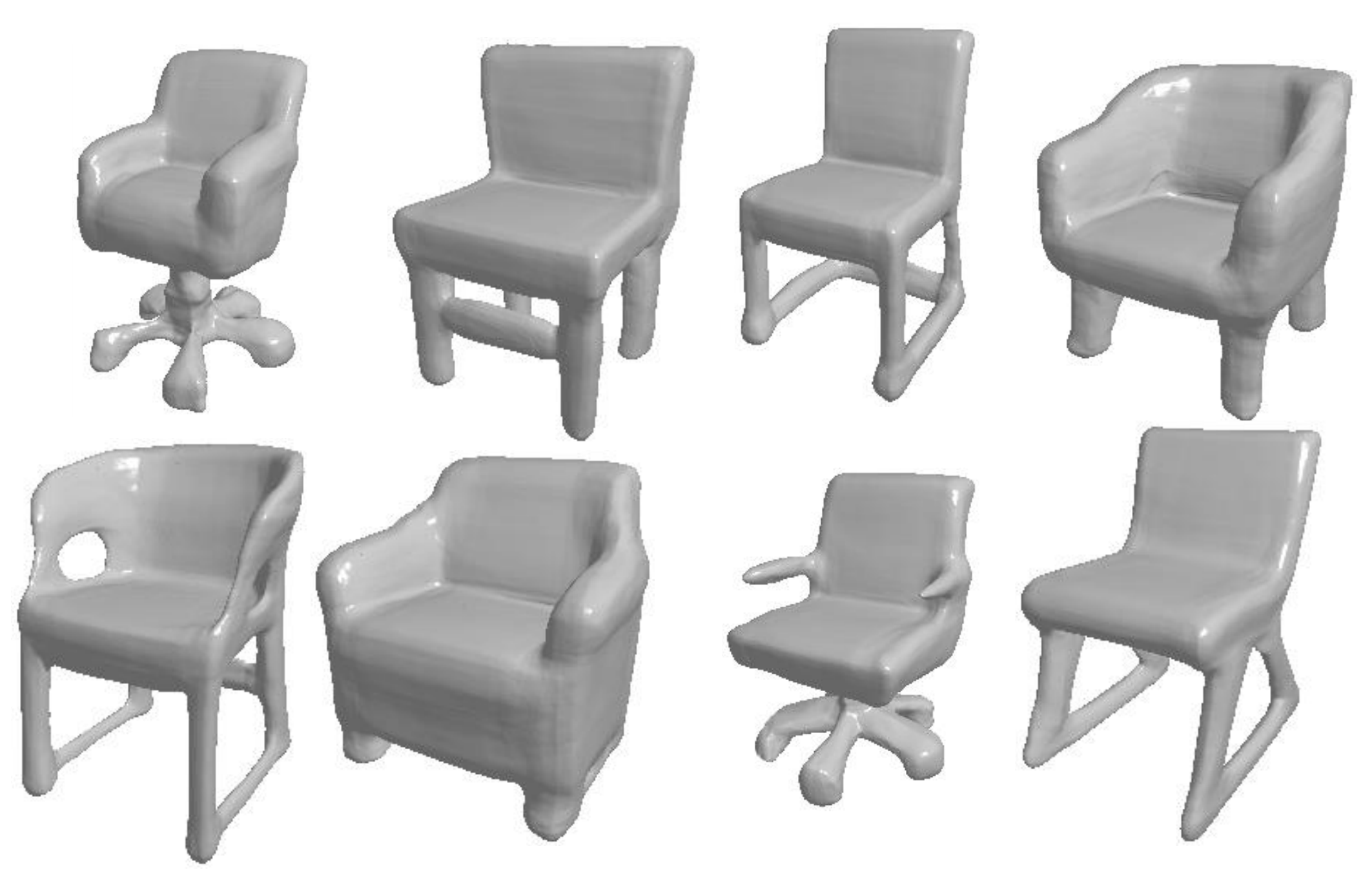}

  \caption{Examples of chairs with complex structure generated by our model trained on ShapeNet~\cite{chang2015shapenet} chair class.}
  \vspace{-0.8em}
  \label{fig:teaser}
\end{figure}
A major barrier in training a neural network on object surfaces is the inherent irregularity and discrete nature of 3D surfaces. Different objects within the same object class can have vastly different topologies, and can be characterized by different internal connectivity when represented as a triangle mesh. Our key insight is that a discriminator applied to the surface should focus on the geometric properties of the surface (curvature, orientation, topology, etc.), while ignoring properties that are irrelevant to the shape. Here we propose to transform object surface meshes to spherical maps computed with a differentiable function $f_{\mathcal{M} \rightarrow S}$. A spherical map is a surface representation defined on discrete samples from the surface of a sphere. The values at each pixel represents a minimal distance of the object surface along a ray, as well as the surface occupancy along the ray. The spherical map is appealing because it is a singular representation that captures the surface geometry of a shape. Given a spherical map, we then utilize a network with spherical convolution layers to complete our discriminator. 

We propose SurfGen, an end-to-end generative model of 3D shapes that is trained with a discriminator which operates on the explicit zero isosurface of an implicit shape function. We demonstrate that applying adversarial training on the surface of an object leads to generated highly realistic shapes. This results in an approach that can generate high quality shapes with arbitrary topology and resolution. Examples of shapes generated by our model are shown in Figure~\ref{fig:teaser}.

\vspace{0.2em}
\noindent In summary our contributions are three fold:
\vspace{-0.7em}
\begin{itemize}
  \item We introduce a spherical projection operator that takes as input an explicit triangle mesh, and is fully differentiable w.r.t. the vertices.
  \vspace{-0.7em}
  \item We propose SurfGen, an end-to-end differentiable 3D shape synthesis framework which applies an adversarial objective on the zero isosurface of the generator.
  \vspace{-0.7em}
  \item We demonstrate our model can synthesize realistic, high quality shapes that have diverse topology.
\end{itemize}
\section{Related Work}
Our method is related to prior work on learning statistical models for 3D shape analysis and generation. In this section we will discuss deep learning based models for 3D shapes, spherical projections, and differentiable rendering.
\begin{figure*}[t!]
  \centering
  \includegraphics[width=0.8\linewidth]{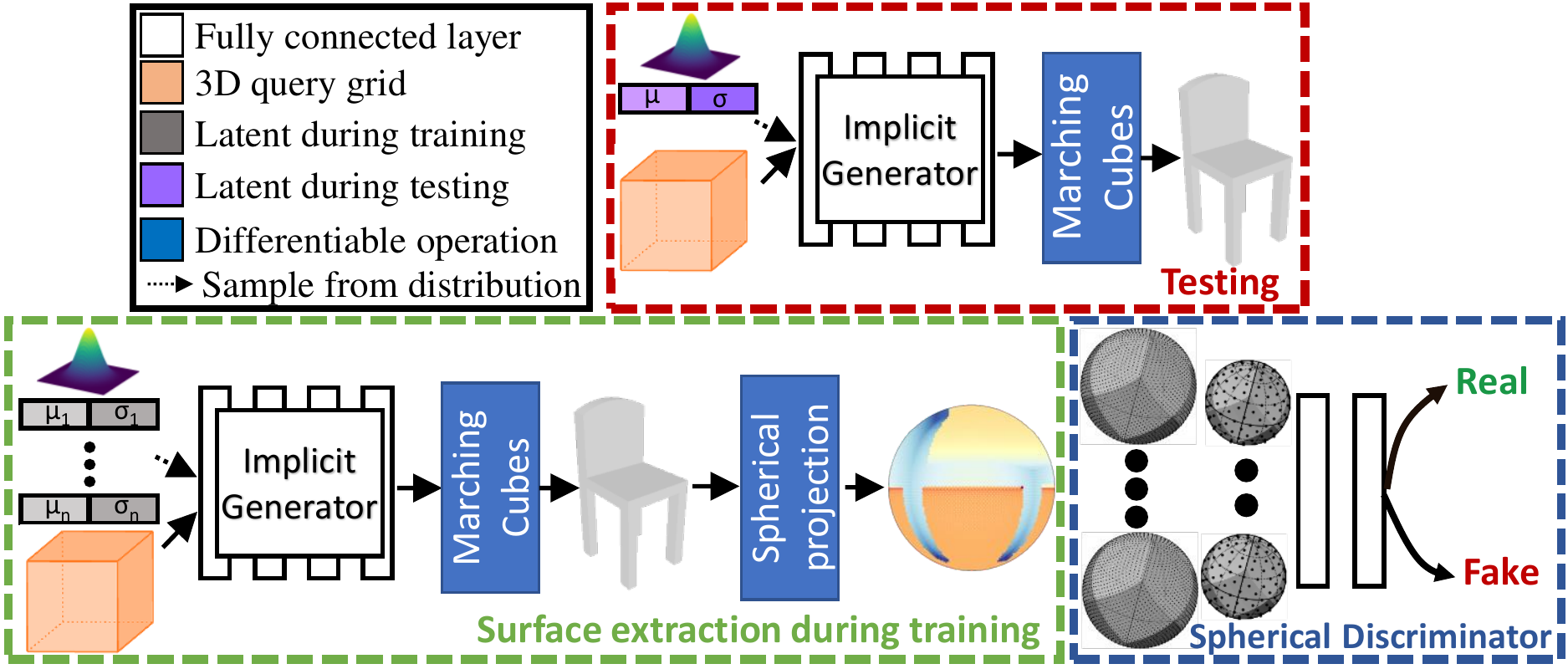}

  \caption{Components of our model.
  \textbf{Top}: During test time, we sample a random code and extract the triangle mesh; \textbf{Bottom}: During training, we extract the surface of the shape represented as a triangle mesh, and compute the differentiable spherical projection of the surface. The spherical surface map is fed to a spherical discriminator. Note that we do not utilize an encoder during training.}
  \vspace{-0.2em}
  \label{fig:net}
\end{figure*}
\myparagraph{Deep learning for 3D shapes.} Recent advances in deep learning have enabled models capable of image-guided shape reconstruction and novel shape synthesis. 3D-R2N2 \cite{choy20163d} proposed learning a recurrent model for voxel based reconstruction with multiple views, \cite{girdhar2016learning, hane2017hierarchical, wu2017marrnet, zhang2018learning} further improved image based voxel reconstruction. Due to memory constraints, these methods are usually limited in terms of their resolution. Methods have also been proposed to generate point clouds from images \cite{fan2017point, wang2019mvpnet}, however the unordered nature of point clouds limits the fidelity of the final 3D reconstructions. Some approaches have been proposed to deform a mesh template \cite{kar2015category, wang2018pixel2mesh, smith2019geometrics}, or a set of surface patches \cite{groueix2018papier}. These techniques are typically limited to modeling shapes of fixed topology, or produce meshes that require post-processing to become usable. It has become possible recently to regress \textit{implicit functions} in the form of binary occupancy or signed distances \cite{xu2019disn, chen2019learning, mescheder2019occupancy, park2019deepsdf}. Similarly, other forms of learned 3D priors have been applied to convert or refine existing shape representations \cite{williams2019deep, badki2020meshlet, erler2020points2surf}. 

\noindent \myparagraph{Generative models for 3D shapes.} Modern generative models for 3D shapes generally utilize an adversarial constraint, take a variational approach, or use flow-based models. Of note, \cite{wu20153d} design a deep belief network to synthesize novel shapes when trained on a large-scale dataset. \cite{wu2016learning} proposed using a generative adversarial network (GAN) for voxels. \cite{achlioptas2018learning, zamorski2018adversarial} proposes learning latent-GANs~\cite{makhzani2015adversarial} for point clouds, while \cite{li2018point} proposes an autoencoding objective to stabilize the training of point cloud generative models. Similarly \cite{gadelha2018multiresolution, valsesia2018learning, shu20193d} advocate for modified generators to improve point cloud generation. Implicit generative models based on latent-GANs \cite{chen2019learning}, and point cloud discriminators \cite{kleineberg2020adversarial} have also been proposed. While these methods are capable of generating high resolution shapes, they still struggle with thin structures and implausible objects.
\myparagraph{Spherical representations.} A spherical representation captures information about a shape on the surface of a sphere. For the purpose of shape retrieval, \cite{ankerst19993d, kazhdan2003rotation, vranic2001tools} represent shapes as spherical distance functions, \cite{yu20033d} captures the number of surface intersections. Spherical representations have also been used in modern deep learning systems for object recognition~\cite{cao20173d}. ~\cite{zhang2018learning} uses a non-differentiable spherical representation computed from voxel grids to facilitate 3D shape reconstruction in the projected spherical space, demonstrating the effectiveness of spherical projections for shape synthesis. In our work, we propose a fully differentiable spherical projection to capture the statistics of surfaces, allowing for the end-to-end training of our generator.

\myparagraph{Differentiable rendering. } 
A spherical projection can be viewed as a rendering operation using a non-linear projection operator. Because the rendering operation is normally discrete, it does not provide usable error gradients for optimization. A variety of mesh~\cite{de2011model, loper2014opendr, li2018differentiable, kato2018neural, liu2019soft, chen2019learning, ravi2020accelerating}, point-cloud, and implicit~\cite{liu2020dist,niemeyer2020differentiable, mildenhall2020nerf, sitzmann2019scene} based differentiable renderers have been proposed. 
\cite{kato2018neural} developed an approximation of gradient for rasterization. \cite{tewari2018self} proposes learning high frequency faces details using a differentiable renderer. \cite{luo2020end} also demonstrated the feasibility of differentiable rendering in 3D scene optimization. \cite{yan2016perspective} proposed a differentiable volume sampling method which can approximate the ray tracing algorithm. Other works such as \cite{kanazawa2018learning, henderson2018learning} also implemented differentiable operation for single image 3D shape reconstruction tasks. Our work enables gradient based optimization in a spherical projection layer, allowing a generator to be optimized by error signals defined in the spherical domain.

\section{Approach}

We propose \textbf{SurfGen}, an end-to-end fully differentiable framework for the generation of 3D shapes. Key to our model, is a differentiable spherical projection operator which allows surfaces to be represented as a spherical projection map, where an adversarial loss can be naturally applied. Our model is illustrated in Figure~\ref{fig:net}.

\subsection{3D Shape Generator}\label{generator}
We adopt DeepSDF~\cite{park2019deepsdf} as the generator in our model. In this generator, each shape is represented by an implicit signed distance function (SDF). For each point $\bold{p}=(p_x, p_y, p_z)$ and a given shape $s$, the SDF encodes the distance of the point to the nearest surface: $\text{SDF}_s(\bold{p})=d, d \in \mathcal{R}$. The sign of $d$ encodes if the point is inside (negative) or outside (outside) a given shape. Our generator $g_\psi$ is trained to map a randomly sampled latent code $z$ and a position $p$ to a corresponding SDF value:
\begin{align}
& g_\psi(z, \bold{p}) \approx \text{SDF}_{s}(\bold{p})
\end{align}
The surface of a shape is implicitly represented by the zero isosurface of the SDF.

\subsection{Differentiable triangle mesh extraction}
In order to apply a discriminator to the surface of an object represented as an implicit SDF, we need to first find the zero isosurface. We choose to utilize marching cubes~\cite{lorensen1987marching} to extract an explicit triangle mesh from the signed distance function. While sphere tracing can also be utilized to extract the zero isosurface from a signed distance function, we found the required speed-accuracy trade-off unsuitable for use in training. 

Instead, we use the MeshSDF method proposed in \cite{remelli2020meshsdf} for differentiable isosurface extraction. During training, we evaluate our signed distance function generator $g_\psi$ on an euclidean grid of size $128^3$ in the range of $[-1,1]$, and use marching cubes (MC) to extract the surface as a triangle mesh $\mathcal{M}=(V,F)$, where $V = \{v_j\}_{j=1}^{M}$ is the set of mesh vertices represented in $\mathbb{R}^3$, and $F$ represents the set of triangle faces enclosed by the edges.

We use the loss from a discriminator $D_\phi$ to compute a gradient with respect to the vertices $V$ from the zero isosurface. For a discriminator loss $\mathcal{L}$ differentiable w.r.t. vertices $V$, the gradient with respect to generator weights $\psi$ can be computed by evaluating:
\vspace{-0.5em}
\begin{align}
&n(v) =\nabla g_\psi(v, z)  \text{ for } v \in V\\
&\frac{\partial \mathcal{L}}{\partial \psi} = -\frac{\partial \mathcal{L}}{\partial v} \cdot n(v) \text{ for } v \in V
\label{eq:meshsdf}
\end{align}
This approach to differentiable surface extraction allows discriminator gradients present on the vertices $V$ to modify surface shape and topology by changing the underlying signed distance function. We refer readers to \cite{remelli2020meshsdf} for a proof and discussion of this method.

\subsection{Differentiable spherical surface projection}
In this section, we present our diffentiable spherical projection layer. Consider a 3D object parameterized as a triangle mesh $\mathcal{M} = (V, F)$. Because surfaces themselves can vary in topology, and meshes can themselves vary in internal connectivity, it is non-trivial to train a neural network on the surface on an object. 

Our spherical projection layer $f_{\mathcal{M} \rightarrow S}$ transforms an irregular 3D mesh into a regular spherical domain. The support of the spherical representation is defined as discrete samples on the unit sphere $S^2$ with $\theta \in [0, 2\pi]$ and $\phi \in [0, \pi]$. 

\begin{figure*}[t!]
  \centering
  \includegraphics[width=0.75\linewidth]{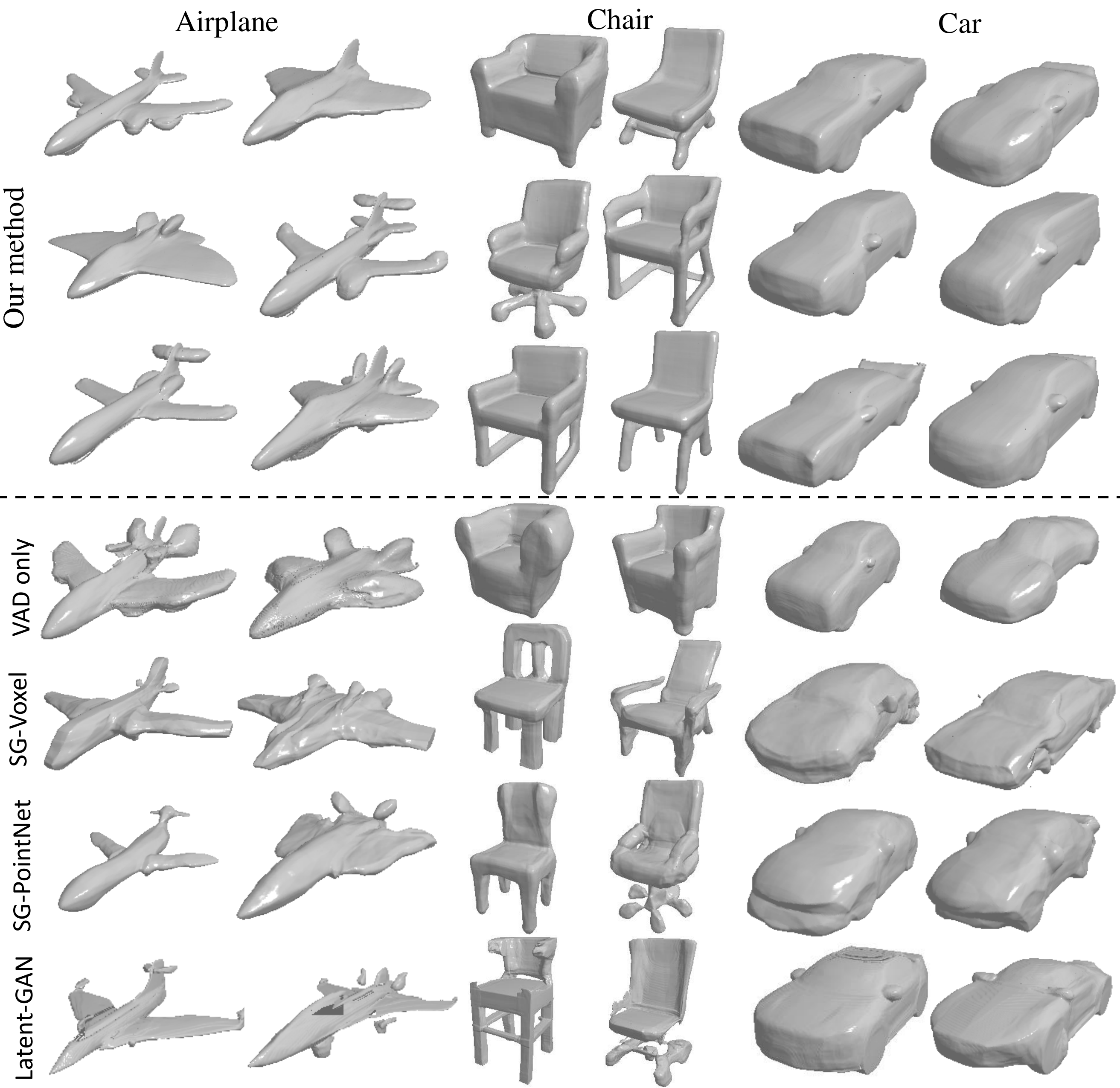}

  \caption{Qualitative results for shapes generated by our SurfGen model; In comparison with samples from other works evaluated on the same testing set: VAD loss only~\cite{zadeh2019variational}, ShapeGAN w/ voxel~\cite{kleineberg2020adversarial} discriminator, ShapeGAN w/ PointNet discriminator~\cite{kleineberg2020adversarial}, and DeepSDF with a latent generator~\cite{achlioptas2018learning,chen2019learning}.}
  \vspace{-1.5em}
  \label{fig:main}
\end{figure*}
There are two major challenges to adopting such a framework. First, the pixel coordinates lie on the surface of a sphere parameterized by $(\theta, \phi)$, which prevents the use of a projection matrix to express the vertex transformations. Second, the rasterization and z-buffer algorithm are discrete operations, which causes discontinuities in the back-propagated gradients as triangles change in depth or move laterally. We use a similar approach to those presented in~\cite{liu2019soft, chen2019dibrender} to enable a differentiable spherical surface projection function.

Because of the non-linear projection, we utilize ray-casting to find intersections of the object surface with rays that originate from the unit sphere. Each ray $R_i$ is defined as a six tuple representing origin and direction in euclidean coordinates:
\begin{align}
   \overrightarrow{R_i} = (O_x, O_y, O_z, D_x, D_y, D_z)_i
\end{align}

We modify the ray-intersection kernel to output direct intersections, as well as "near misses" where the ray is within distance $r$ of a triangle. The ray-intersection kernel is repeated such that the $k$ nearest hits along each ray are retrieved:
\begin{align}
{(\overrightarrow{p_{j^{i}}}, F_j)}_{j=1}^{k} = \texttt{rayintersect}_{k,r}(\mathcal{M}, R_i)
\end{align}
\begin{figure}[t!]

  \includegraphics[width=\linewidth]{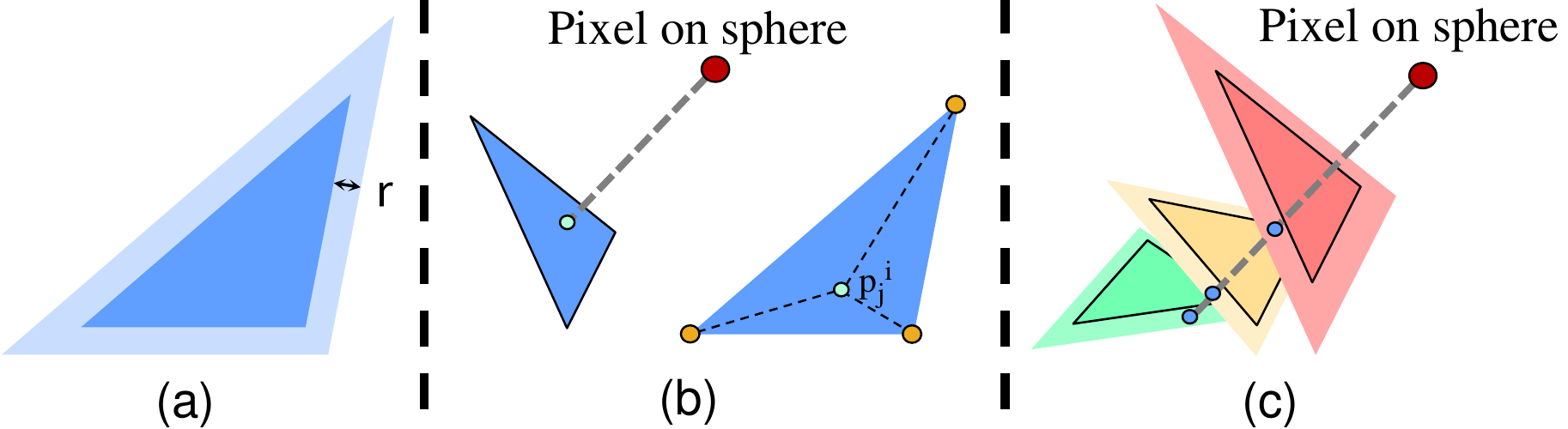}
  \vspace{-1.5em}  
  \caption{ Differentiable spherical projection from the perspective of a ray. \textbf{(a)} The search region of each ray includes the triangle (solid) and a small "radius" $r$ neighborhood (transparent). \textbf{(b)} Direct hits inside a triangle allows us to interpolate depth. \textbf{(c)} "Near miss" hits in the neighborhood of a triangle allows for occupancy to be computed.}
  \vspace{-1.3em}
  \label{fig:ray}

\end{figure}

At an intersection location $\overrightarrow{p_{j^{i}}}$ inside triangle $j$ for $R_i$, we compute a pixel attribute $u^i_j$ as an barycentric interpolation of vertex attributes

\begin{align}
u^{i}_j &= w^i_{j,0} \cdot u_{j,0} + w^i_{j,1} \cdot u_{j,1} + w^i_{j,2} \cdot u_{j,2}\\
w^{i}_{j,q} &= \Omega_{q}(\overrightarrow{p_{j^i}}, \overrightarrow{v_{j,0}}, \overrightarrow{v_{j,1}},\overrightarrow{v_{j,2}}); q \in \{0, 1, 2\}
\end{align}
where the barycentric weights are computed using the differentiable function $\Omega$. When computing a spherical depth projection, we set the attributes to be the $(x,y,z)$ coordinate of each vertex in triangle $j$:
\begin{align}
(u_{j,0},u_{j,1},u_{j,1}) = (v_{j,0},v_{j,1},v_{j,2})
\end{align}

Beyond the spherical depth map, the spherical silhouette of a shape can also be computed such that it is differentiable. We utilize "near miss" rays that do not intersect inside a triangle, but has a point on that ray $p^{'}_{j^{i}}$ that is within distance $r$ outside of a triangle. We compute the squared euclidean distance between the ray and the closest point on the triangle:
\begin{align}
&d(p^{'}_{j^{i}}, F_j) = \min_{p \in F_j}\norm{p-p^{'}_{j^{i}}}^{2}_{2}\\
&\alpha_j^{i} = \text{exp}{\frac{-d(p^{'}_{j^{i}}, F_j)}{\delta}}
\end{align}

Where $\delta$ is a hyper-parameter that controls how fast the influence decays as a function of distance. We use a product based function to aggregate all the near miss collisions for a given ray:
\begin{align}
\alpha^i = 1 - \prod_{j}(1-\alpha^{i}_j)
\end{align}

A ray is illustrated in Figure~\ref{fig:ray}. This allows for the efficient computation of spherical depth and occupancy maps in a way that is fully differentiable with respect to the vertex attributes.

\subsection{Choice of sampling and projection}

\myparagraph{Sampling on a Sphere.} A discretization of the sphere needs to be selected prior to spherical projection. Common sampling schemes includes Driscoll and Healy~\cite{driscoll1994computing}, icosahedral~\cite{wenninger1999spherical}, and HEALPix~\cite{gorski2005healpix}; which use rectangular, icosahedral, and equal area sampling respectively. Due to our efficient ray-casting based implementation, the differentiable spherical projection layer can work with any discretization of the sphere. In practice, the HEALPix based sampling with $12,288$ points is used due to the balanced area assigned to each point on the sphere. 

\myparagraph{Choice of Projection.} \label{projections}
\begin{figure}[t!]
  \centering
  \includegraphics[width=0.9\linewidth]{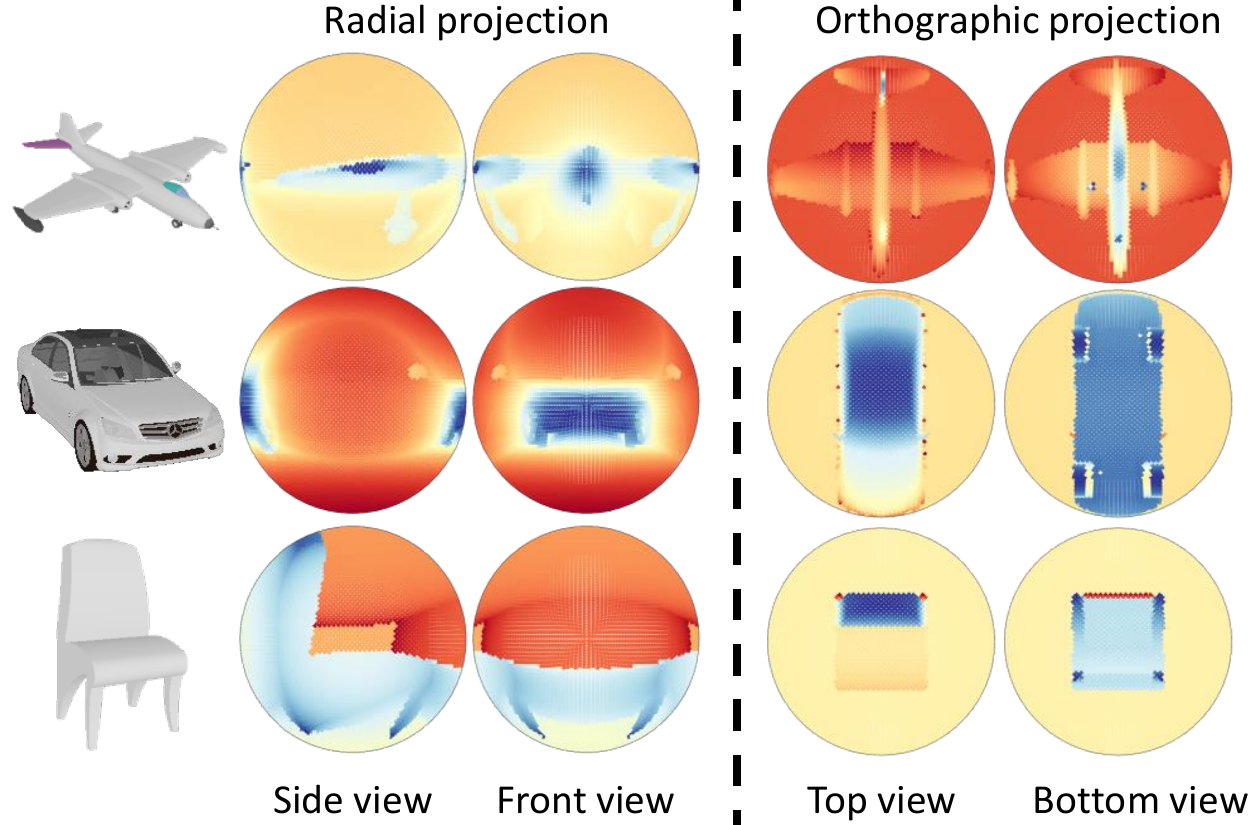}

  \caption{Visualization of radial and orthographic projections. These depth projections reflect the surface distance from the origin of the ray.}
  \label{fig:radialsph}

\end{figure}
Our spherical projection layer can model arbitrary projections. The most simple projection operator is the \textit{radial spherical projection}, which gives a ray direction to be $(D_x, D_y, D_z) = -(O_x, O_y, O_z)$, where $O$ is discrete sample on the sphere. It was shown in \cite{zhang2018learning}, that a radial projection may have poor coverage of a shape due to self-occlusions. They address this by combining the spherical representation with an additional non-radial projection. Taking inspiration from orthographic map projections, we propose \textit{orthographic spherical projection} where $(D_x, D_y, D_z) = -(0, 0, O_z)$, where $z$ is assumed to be the gravity aligned (top-down) axis. Our full spherical projection combines the spherical depth map from radial \& orthographic spherical projections, and use the silhouette map from the orthographic projection. The radial silhouette map is omitted since most radial rays have direct intersections within a triangle face, and do not provide a useful training signal. We show how the projections compare in Figure~\ref{fig:radialsph}.

\subsection{Discriminator Implementation}
\myparagraph{Network Architecture.} As our spherical maps are discrete samples on a unit sphere and non-euclidean, we cannot use a regular 2D or 3D convolution. We implement our discriminator using the graph-based spherical convolution layers proposed by \cite{defferrard2020deepsphere}. Our discriminator takes as input the 3 spherical maps described in section~\ref{projections}. The network consists of 5 residual spherical blocks that each perform average pooling to reduce the total number of spherical samples by a factor of 4, as well as a single self-attention layer in the discriminator. We average pool the final 12 pixels, 256 channel spherical maps, and use fully connected layers to produce the scalar discriminator output. We utilize Instance Normalization~\cite{ulyanov2016instance} and Leaky ReLU~\cite{maas2013rectifier} in the discriminator.

\subsection{Overall model} The straightfoward approach to training an implicit generator with a surface discriminator would be to update the implicit generator and surface discriminator in an alternate fashion. However, unlike the SDF which is defined for all $p \in \mathbb{R}^3$, the surface only exists at select locations. 
\begin{figure*}[t!]
  \centering
  \includegraphics[width=0.7\linewidth]{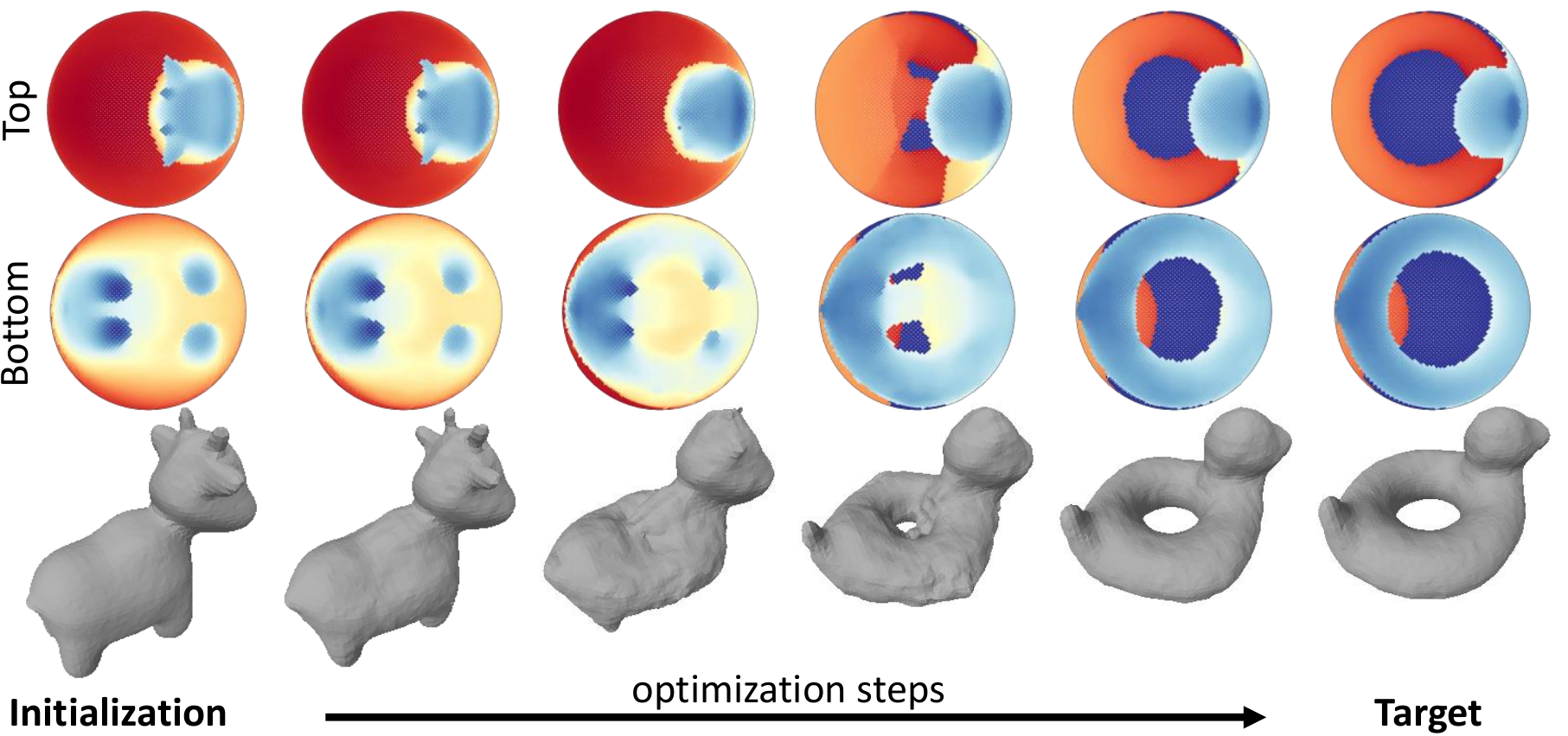}
  \vspace{-0.2em}
  \caption{We optimize a MSE loss on the spherical representation between the target and initial shape. \textbf{From top to bottom:} the top view of the radial projection, the bottom view of the radial projection, rendered SDF output. This shows that our spherical surface projection can induce topology and shape changes.}
  \vspace{-0.5em}
  \label{fig:optimization}
\end{figure*}
We need to constrain the generator such that the location of the isosurface is within the spherical projection layer, and ideally approximate to the desired shape. Methods using geometric initializations~\cite{atzmon2020sal}, meta-learning~\cite{sitzmann2020metasdf}, and variational methods~\cite{hao2020dualsdf} have been proposed to stabilize training and accelerate convergence. We choose to regularize our generator with a variational autodecoder (VAD)~\cite{hao2020dualsdf} loss alongside our adversarial criterion. This VAD-GAN setup preserves the generator$+$discriminator structure of a generative adversarial network (GAN) during training, only requiring the addition of a light-weight embedding layer to the model, while significantly outperforming the original VAD only objective. 

During training, we sample a latent code $z$ from the approximate posterior for a given shape $x_j$ modeled as a multivariate Gaussian with diagonal covariance:
\begin{align}
    q(z | x_j) := \mathcal{N}(z; \mu_{j}, \sigma^2_j\cdot\mathbb{I})
\end{align}
 We use the reparameterization trick~\cite{kingma2013auto} to allow for direct optimization of latent parameters $\mu_{j}, \sigma_{j}$, and set $z_j = \mu_j + \sigma_j \odot \epsilon$, where $\epsilon \sim \mathcal{N}(0, \mathbb{I})$. During training, the generator and latent parameters are trained to minimize the the adversarial objective $\mathcal{L}_{GAN}$ and maximize the evidence lower bound of the marginal likelihood (\textbf{ELBO}):
\begin{align}
\log p_\theta(x) &\geq \underset{{z\sim q(z|x)}}{\mathbb{E}}\biggl[\log p(x|z)\biggr] - D_{KL}(q(z|x)||p(z))\notag
\end{align}
\vspace{-3mm}
\begin{align}
&p(x|z) = -\underset{{\bold{p}}}{\mathbb{E}}\biggl[\mathcal{L}_{SDF} \biggl(g_\psi(z, \bold{p}), \text{SDF}_s(\bold{p}))\biggr)\biggl]\label{eq:px_g_z}
\end{align}
Equation~\ref{eq:px_g_z} could be approximated by sampling from 3D space following a certain distribution defined in~\cite{park2019deepsdf}. For the final results, we use clamped version of L1 loss for $\mathcal{L}_{SDF}(\cdot)$.
To stabilize GAN training, we employ standard hinge GAN loss $\mathcal{L}_{\text{hinge}}$~\cite{lim2017geometric, tran2017deep} with a small amount of feature matching loss $\mathcal{L}_{\text{feat}}$~\cite{salimans2016improved} applied to the first four discriminator feature maps. 

In summary, the overall objective is summarized in Equation~\ref{eq:overall_obj} where we use $\alpha=1$, $\beta=1$ and $\gamma=1e^{-5}$, $\lambda=0.5$.
\begin{align}
    \mathcal{L}_{GAN} &= \mathcal{L}_{\text{hinge}} + \lambda\mathcal{L}_{\text{feat}} \\
    \mathcal{L}_{\text{total}} &= \alpha\mathcal{L}_{GAN} + \beta \mathcal{L}_{SDF} + \gamma D_{KL}\label{eq:overall_obj}
\end{align}

In order to facilitate stable training, we zero the gradients in the generated depth maps where there is no surface occupancy in the corresponding target shape:
\begin{align}
    \text{Depth}_\text{grad}[O_\text{target}<1.0] = 0
\end{align}

To ensure that surface gradients w.r.t. the SDF generator are within the same magnitude as the other losses, we scale the gradient from Equation~\ref{eq:meshsdf} with a constant $\omega=1e^{-4}$.

\begin{table}[]
\resizebox{\linewidth}{!}{%
\begin{tabular}{l|lll|lll}
\hline
                  & \multicolumn{3}{c|}{Chamfer}                        & \multicolumn{3}{c}{EMD}                             \\ \hline
Iterations        & 5               & 10              & 30              & 5               & 10              & 30              \\ \hline
Radial Depth      & 0.0486          & \textbf{0.0327} & 0.0032          & 0.0389          & 0.0244          & 0.0029          \\
Ortho Depth       & 0.0496          & 0.0335          & 0.0024          & 0.0384          & 0.0234          & 0.0026          \\
Radial silhouette & 0.0584          & 0.0590          & 0.0587          & 0.0446          & 0.0443          & 0.0433          \\
Ortho silhouette  & 0.0556          & 0.0528          & 0.0460          & 0.0443          & 0.0373          & 0.0322          \\
Combined          & \textbf{0.0478} & 0.0329          & \textbf{0.0022} & \textbf{0.0356} & \textbf{0.0219} & \textbf{0.0010} \\ \hline
\end{tabular}%
}
\vspace{0.2em}
\caption{We compare the effectiveness of four spherical features in terms of Chamfer and earth movers distance (EMD) during the optimization process. \textit{Combined} indicates the radial depth, ortho depth, and ortho silhouette are used together. \textbf{Bold} indicates best method at each iteration.}
\label{table:opti}
\end{table}

Our model is implemented using Pytorch, with the ray casting implemented in embree~\cite{wald2014embree, samuel_f_potter_2021_4609402}, and marching cubes implemented in CUDA~\cite{mcubes_pytorch}.
\section{Experiments}
    \begin{table*}[t!]\centering
\ra{0.7}
\begin{tabular}{p{1.5cm}p{2.9cm}p{2.5cm}p{1.0cm}p{0.1cm}p{1.0cm}p{1.0cm}p{0.1cm}p{1.0cm}p{1.0cm}}\midrule
& & && \phantom{a}& \multicolumn{2}{c}{MMD~($\downarrow$)} & \phantom{a}& \multicolumn{2}{c} {COV~(\%, $\uparrow$)}\\
\cmidrule{6-7} \cmidrule{9-10}
Category& Model & Discriminator &JSD~($\downarrow$)&& CD & EMD && CD & EMD\\ \midrule

\multirow{5}{*}{Airplane} &ShapeGAN-Voxel & Voxel &0.2848&& 0.0193 & 0.1818 && 0.0794 & 0.0918 \\

& ShapeGAN-PN & PointNet&0.2393 && 0.0119 & 0.1670 && 0.1066 & 0.1191\\

& Latent-GAN & Latent &0.3643 && 0.0138 & 0.1871 && 0.0843 & 0.0893\\
& VAD-SDF & None&0.2221 && 0.0103 & 0.1500 && 0.1141 & 0.1190\\

& \textbf{SurfGen}~(ours) & Surface &\textbf{0.1586}&& \textbf{0.0074} & \textbf{0.1371} && \textbf{0.1191} & \textbf{0.1215}\\

\midrule

\multirow{5}{*}{Chair} &ShapeGAN-Voxel & Voxel & 0.0307&& 0.0231 & 0.2042 && 0.3408 & 0.3537\\

& ShapeGAN-PN & PointNet &0.0304&& 0.0152 & 0.1711 && 0.3641 & \textbf{0.3868}\\

& Latent-GAN & Latent&0.0394 && 0.0125 & 0.1660 && 0.3742 & 0.3667\\
& VAD-SDF & None&0.0846 && 0.0142 & 0.1662 && 0.2051 & 0.2180\\

& \textbf{SurfGen}~(ours) & Surface &\textbf{0.0287}&& \textbf{0.0095} & \textbf{0.1440} && \textbf{0.3812} & 0.3586\\

\midrule

\multirow{5}{*}{Car} &ShapeGAN-Voxel & Voxel &0.0336&& 0.0056 & 0.1221 && 0.3181 & 0.2869 \\
& ShapeGAN-PN & PointNet &\textbf{0.0259}&& 0.0051 & 0.1061 && \textbf{0.3409} & \textbf{0.3579} \\
& Latent-GAN & Latent &0.0649&& 0.0061 & 0.1292 && 0.2784 & 0.2556\\
& VAD-SDF & None &0.0568&& 0.0048 & 0.1063 && 0.2414 & 0.2585\\
& \textbf{SurfGen}~(ours) & Surface &0.0463&& \textbf{0.0038} & \textbf{0.0982} && 0.2755 & 0.3267\\

\midrule
\end{tabular}
\caption{Generation results across ShapeNet classes; $\downarrow$ indicates that a lower value is better, $\uparrow$ indicates a higher value is better. }
\label{table:main_compare}
\end{table*}

\subsection{Validating end-to-end optimization}

We first verify the differentiability of our spherical projection layer. Using the data from~\cite{remelli2020meshsdf}, we train a generator network $g_\psi$ to approximate the signed distance function of two shapes: a genus zero cow, and a genus one rubber duck. The generator learns to associate a latent code with a corresponding implicit distance field. The loss during optimization is the pixel-wise mean squared error (MSE) between the current and target spherical projections:
\begin{align}
    \frac{1}{N}(f_{\mathcal{M} \rightarrow S}(g_\psi(z))-S_\text{target})^2
\end{align}
The gradients are backpropagated and are used to optimize the latent code $z$. As shown in Figure~\ref{fig:optimization}, our spherical projection layer can modify the underlying implicit representation, and can change both the shape and topology of an object.

We quantitatively compare the effect of using radial depth, radial silhouette, orthographic depth, orthographic silhouette, as well as combined (radial depth + orthographic depth + orthographic silhouette) as features for our optimization in Table~\ref{table:opti}. Because few rays in the radial projection experience a "near miss" for the shape, the radial silhouette does not provide useful gradients for the optimization process. The loss that uses the three combined features provides the fastest convergence when measured by chamfer distance or earth movers distance. The final projection used in our shape generation experiment uses the combined features.

\subsection{Shape Generation}
\myparagraph{Data preparation.}\noindent All models are trained on one of three categories from ShapeNet.v2~\cite{chang2015shapenet}: \textit{airplane}, \textit{car}, and \textit{chair}. To ensure comparability, we use the official training split.

Each shape is centered and normalized to the unit sphere. We utilize the improved signed distance generation method proposed by the authors of ShapeGAN. Each shape is rendered from 50 equidistance views, the depth buffer is projected into object space to compute the surface point cloud. A point is considered to be outside the shape if it is seen by any camera. Shapes are discarded if fewer than $0.5\%$ of the points are inside. We apply a small negative offset ($2e^{-3}$) to SDF values during training to facilitate surface extraction. This results in $2788$, $2452$, $4550$ shapes in the training set for the \textit{airplane}, \textit{car}, and \textit{chair} categories respectively. We use the same sampling scheme used in DeepSDF. For the ground truth test set, we uniformly query points in a unit sphere, and randomly select $2048$ points thresholded to lie near the surface. 
\begin{figure}[]
  \centering
  \includegraphics[width=0.8\linewidth]{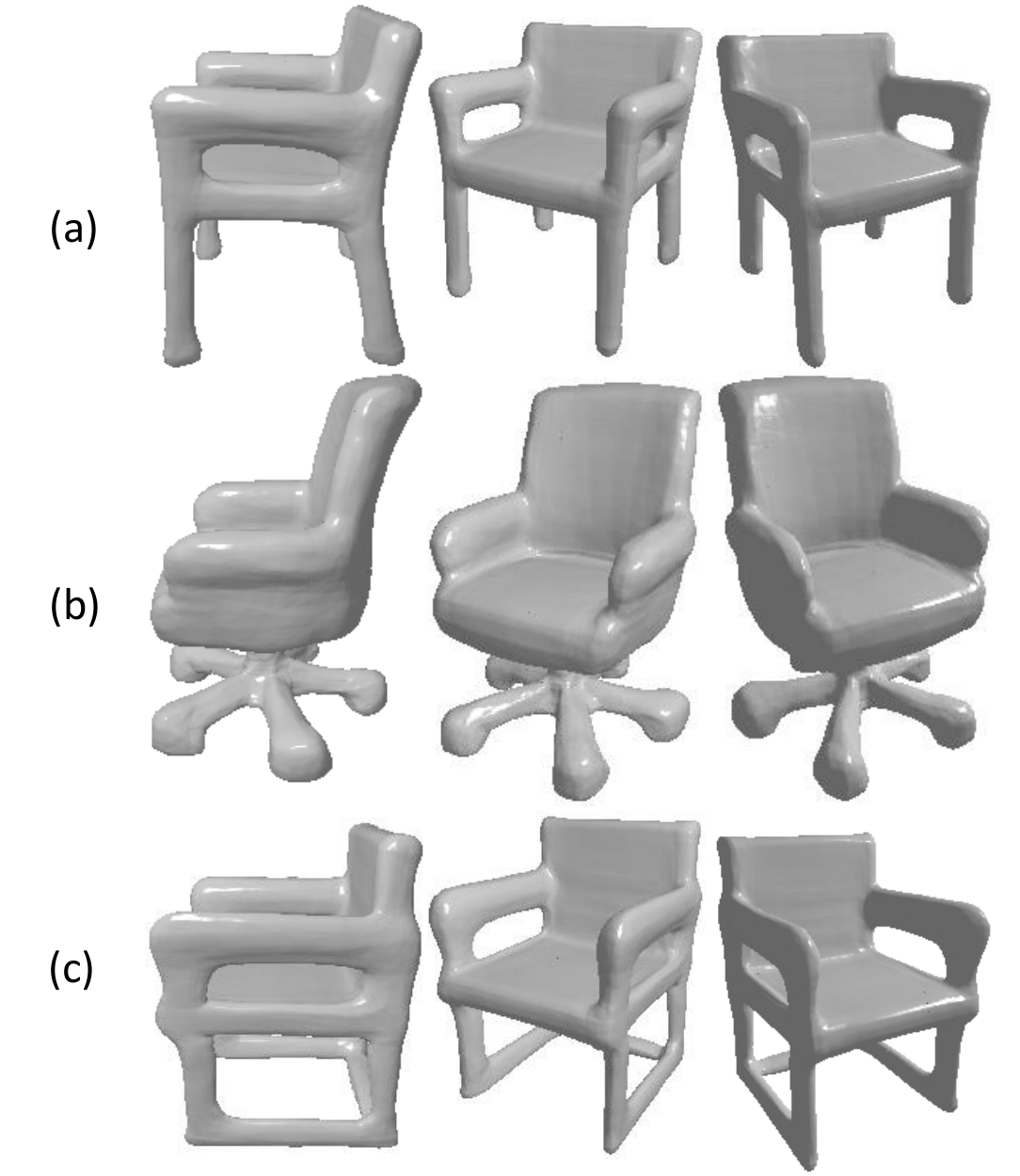}

  \caption{Multiple views of synthesized shapes by our model.}
  \vspace{-1.0em}
  \label{fig:multi}

\end{figure}

\myparagraph{Baselines.} \noindent We compare against four alternative models for implicit shape generation: two variants of ShapeGAN~\cite{kleineberg2020adversarial} which use a voxel and PointNet as discriminator, a latent-GAN trained on DeepSDF embeddings, and a VAD based SDF model which does not utilize our surface discriminator. We use the VAD layer implementation provided in~\cite{hao2020dualsdf}. For the latent-GAN, VAD, and our SurfGen model, we use the DeepSDF network as backbone. For ShapeGAN the author provided code and hyperparameters are used.
\begin{figure}[]
  \centering
  \includegraphics[width=0.8\linewidth]{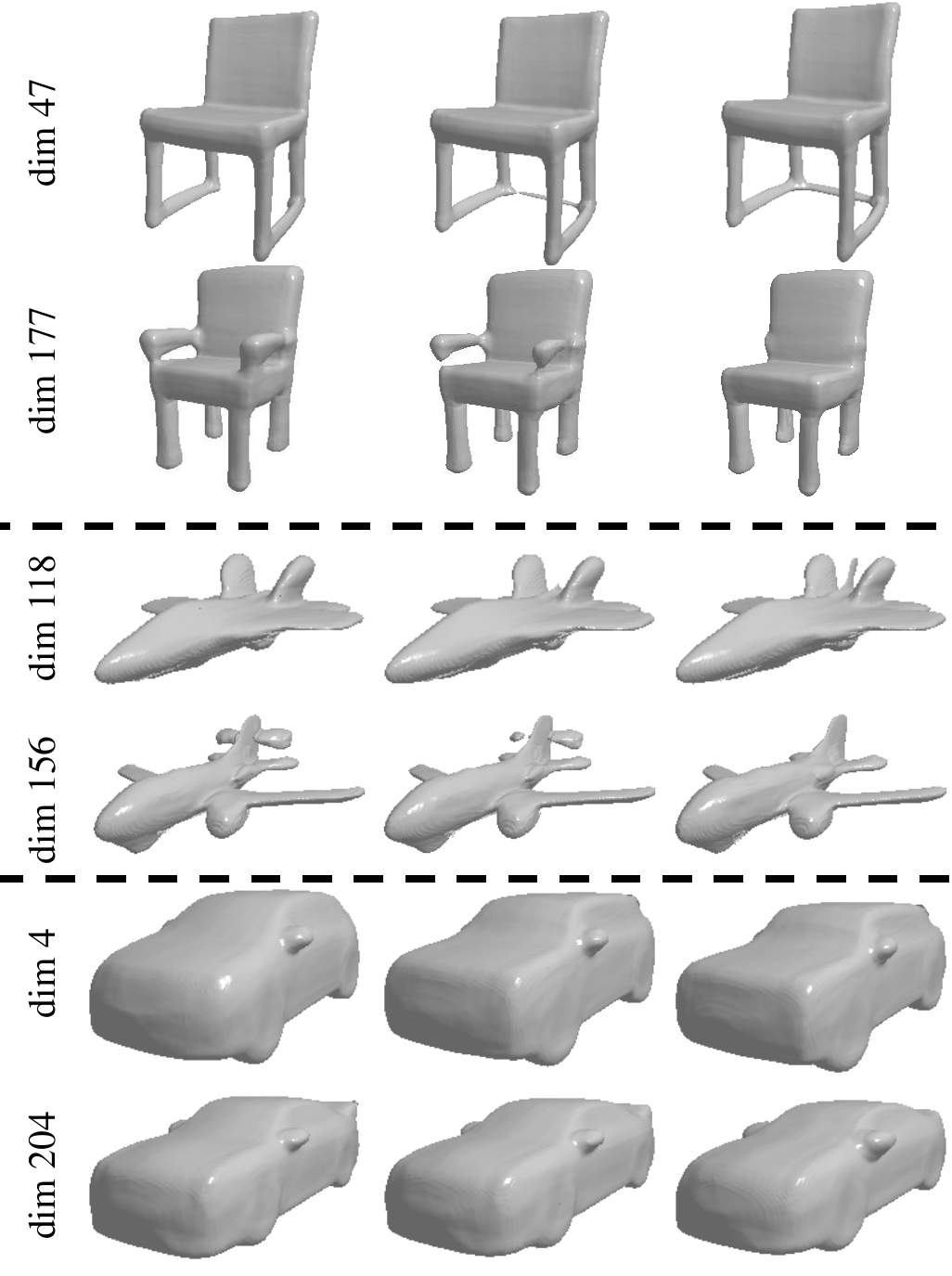}

  \caption{The effect of manipulating individual dimensions of the latent vector}
  \vspace{-0.5em}

  \label{fig:latent_dim}
\end{figure}
\myparagraph{Evaluation and metrics.}
Every mesh is extracted using marching cubes at $256^3$ resolution, and $2048$ points are randomly sampled from each mesh. The results are compared using the metrics introduced by~\cite{achlioptas2018learning}. Minimum matching distance (MMD) measures the distance of a shape from the test set to its nearest neighbor, and can be understood to be a proxy for \textit{fidelity}. Coverage (COV) measures the fraction of the test set that are the nearest neighbor to a sample in the generated set, and can be understood to be a proxy for diversity. For each measure, the distance metric can either be chamfer distance (CD) or earth movers distance (EMD).

\myparagraph{Results.} Qualitative results for shapes generated by each method are shown in Figure~\ref{fig:main} and Figure~\ref{fig:multi}, while quantitative results are shown in  Table~\ref{table:main_compare}. Shapes generated by our model have high-quality surfaces that are largely free from the high frequency artifacts present in Latent-GAN, and the block like artifacts from the two ShapeGAN variants. We also observe many thin shell like artifacts on VAD synthesized airplanes. For the chair class, the VAD only method struggles to capture shapes with complicated geometry, and generally synthesizes chairs with no thin parts.

SurfGen has the lowest MMD value across all evaluated classes for both CD \& EMD, this indicates that our methods synthesizes shapes are closely matched to samples from the test set. SurfGen produces high coverage for the airplane class, while it is tied for coverage in the chair class. Our method is less competitive in the car class. A possible reason is the lack of high frequency geometry normally present on the outside surface of cars. This causes cars to be difficult to learn using a surface based adversarial loss.

\subsection{Latent Analysis}
\myparagraph{Individual latent dimensions.} We explore if individual dimensions in our latent space have semantic meaning. Given a randomly sampled latent code, we select a dimension and increase its value in small increments. Our results are shown in
Figure~\ref{fig:latent_dim}. We observe that changing certain dimensions can induce structural changes in a shape.
\vspace{-1em}
\begin{figure}[]
  \centering
  \includegraphics[width=0.9\linewidth]{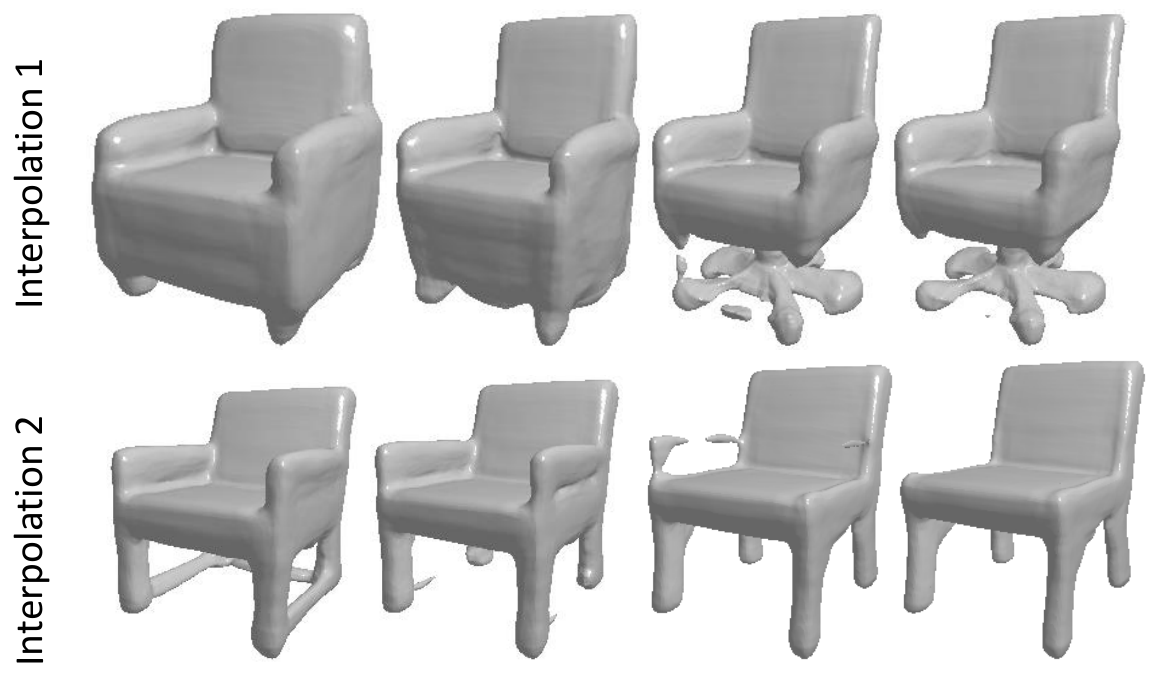}

  \caption{Visualization of the shape as we linearly interpolate between two randomly sampled latent vectors.}
  \label{fig:interp}

\end{figure}
\myparagraph{Interpolating between latents.}\noindent In Figure~\ref{fig:interp}, we demonstrate that shapes can smoothly change as we linearly interpolate between two randomly sampled latent vectors, even between shapes that have different topologies.
\vspace{-1em}
\myparagraph{Computational efficiency.}\noindent For a batch of 8 shapes, on a dual \texttt{Nvidia 3090} system the grid query step at $128^3$ resolution takes approximately $2,500$ ms, while marching cubes takes $90$ ms. The spherical projection takes around $600$ ms on a 12 core CPU. For $G$ and $D$ combined, the forward and backwards pass for a single batch takes around $5$ seconds total.

\subsection{Failure Cases} 
\begin{figure}[]
  \centering
  \includegraphics[width=0.6\linewidth]{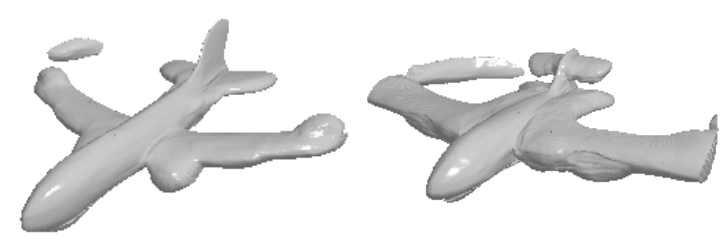}

  \caption{Two failure cases that exhibit ring like artifacts near the wing.}
  \label{fig:fail}
  \vspace{-1em}

\end{figure}
Certain shapes do not have well behaved spherical projections. In the most common case, we observe that this is mostly due to a shape having protrusions along the gravity aligned axis that significantly affect the centering of a shape into a unit sphere. The issue is most prominent in \textit{airplanes}, while this is to a large degree mitigated by using the orthographic projection, we observe rare ring-like artifacts in airplanes as shown in Figure~\ref{fig:fail} that seem to be induced by the spherical discriminator. As part of future work, it may make sense to investigate the integration of the surface discriminator with discriminators directly applied to the implicit field.

\section{Conclusion}
In this paper, we have introduced a novel shape synthesis method that allows a discriminator to focus directly on the surface on a generated shape. Our method is capable of generating diverse high quality shapes with complex geometry. Our model has diverse downstream applications as part of a larger 3D synthesis pipeline. We hope our work will inspire future research in 3D shape synthesis.

\myparagraph{Acknowledgment}
This work was supported by NSF grant CISE RI 1816568, and an interdisciplinary training fellowship in computational neuroscience at Carnegie Mellon University sponsored by NIH NIDA grant 5T90 DA023426. The authors would further like to thank the Chen Institute (TCCI) for their support of this work.
\clearpage
{\small
\bibliographystyle{ieee_fullname}
\bibliography{egbib}

\begin{thebibliography}{10}\itemsep=-1pt

\bibitem{achlioptas2018learning}
Panos Achlioptas, Olga Diamanti, Ioannis Mitliagkas, and Leonidas Guibas.
\newblock Learning representations and generative models for 3d point clouds.
\newblock In {\em International conference on machine learning}, pages 40--49.
  PMLR, 2018.

\bibitem{ankerst19993d}
Mihael Ankerst, Gabi Kastenm{\"u}ller, Hans-Peter Kriegel, and Thomas Seidl.
\newblock 3d shape histograms for similarity search and classification in
  spatial databases.
\newblock In {\em International symposium on spatial databases}, pages
  207--226. Springer, 1999.

\bibitem{atzmon2020sal}
Matan Atzmon and Yaron Lipman.
\newblock Sal: Sign agnostic learning of shapes from raw data.
\newblock In {\em Proceedings of the IEEE/CVF Conference on Computer Vision and
  Pattern Recognition}, pages 2565--2574, 2020.

\bibitem{badki2020meshlet}
Abhishek Badki, Orazio Gallo, Jan Kautz, and Pradeep Sen.
\newblock Meshlet priors for 3d mesh reconstruction.
\newblock In {\em Proceedings of the IEEE/CVF Conference on Computer Vision and
  Pattern Recognition}, pages 2849--2858, 2020.

\bibitem{cao20173d}
Zhangjie Cao, Qixing Huang, and Ramani Karthik.
\newblock 3d object classification via spherical projections.
\newblock In {\em 2017 International Conference on 3D Vision (3DV)}, pages
  566--574. IEEE, 2017.

\bibitem{chang2015shapenet}
Angel~X Chang, Thomas Funkhouser, Leonidas Guibas, Pat Hanrahan, Qixing Huang,
  Zimo Li, Silvio Savarese, Manolis Savva, Shuran Song, Hao Su, et~al.
\newblock Shapenet: An information-rich 3d model repository.
\newblock {\em arXiv preprint arXiv:1512.03012}, 2015.

\bibitem{chen2019dibrender}
Wenzheng Chen, Jun Gao, Huan Ling, Edward Smith, Jaakko Lehtinen, Alec
  Jacobson, and Sanja Fidler.
\newblock Learning to predict 3d objects with an interpolation-based
  differentiable renderer.
\newblock In {\em Advances In Neural Information Processing Systems}, 2019.

\bibitem{chen2019learning}
Zhiqin Chen and Hao Zhang.
\newblock Learning implicit fields for generative shape modeling.
\newblock In {\em Proceedings of the IEEE Conference on Computer Vision and
  Pattern Recognition}, pages 5939--5948, 2019.

\bibitem{choy20163d}
Christopher~B Choy, Danfei Xu, JunYoung Gwak, Kevin Chen, and Silvio Savarese.
\newblock 3d-r2n2: A unified approach for single and multi-view 3d object
  reconstruction.
\newblock In {\em European conference on computer vision}, pages 628--644.
  Springer, 2016.

\bibitem{de2011model}
Martin de La~Gorce, David~J Fleet, and Nikos Paragios.
\newblock Model-based 3d hand pose estimation from monocular video.
\newblock {\em IEEE transactions on pattern analysis and machine intelligence},
  33(9):1793--1805, 2011.

\bibitem{defferrard2020deepsphere}
Micha{\"e}l Defferrard, Martino Milani, Fr{\'e}d{\'e}rick Gusset, and
  Nathana{\"e}l Perraudin.
\newblock Deepsphere: a graph-based spherical cnn.
\newblock {\em arXiv preprint arXiv:2012.15000}, 2020.

\bibitem{driscoll1994computing}
James~R Driscoll and Dennis~M Healy.
\newblock Computing fourier transforms and convolutions on the 2-sphere.
\newblock {\em Advances in applied mathematics}, 15(2):202--250, 1994.

\bibitem{erler2020points2surf}
Philipp Erler, Paul Guerrero, Stefan Ohrhallinger, Niloy~J Mitra, and Michael
  Wimmer.
\newblock Points2surf learning implicit surfaces from point clouds.
\newblock In {\em European Conference on Computer Vision}, pages 108--124.
  Springer, 2020.

\bibitem{fan2017point}
Haoqiang Fan, Hao Su, and Leonidas~J Guibas.
\newblock A point set generation network for 3d object reconstruction from a
  single image.
\newblock In {\em Proceedings of the IEEE conference on computer vision and
  pattern recognition}, pages 605--613, 2017.

\bibitem{gadelha2018multiresolution}
Matheus Gadelha, Rui Wang, and Subhransu Maji.
\newblock Multiresolution tree networks for 3d point cloud processing.
\newblock In {\em Proceedings of the European Conference on Computer Vision
  (ECCV)}, pages 103--118, 2018.

\bibitem{girdhar2016learning}
Rohit Girdhar, David~F Fouhey, Mikel Rodriguez, and Abhinav Gupta.
\newblock Learning a predictable and generative vector representation for
  objects.
\newblock In {\em European Conference on Computer Vision}, pages 484--499.
  Springer, 2016.

\bibitem{gorski2005healpix}
Krzysztof~M Gorski, Eric Hivon, Anthony~J Banday, Benjamin~D Wandelt, Frode~K
  Hansen, Mstvos Reinecke, and Matthia Bartelmann.
\newblock Healpix: A framework for high-resolution discretization and fast
  analysis of data distributed on the sphere.
\newblock {\em The Astrophysical Journal}, 622(2):759, 2005.

\bibitem{groueix2018papier}
Thibault Groueix, Matthew Fisher, Vladimir~G Kim, Bryan~C Russell, and Mathieu
  Aubry.
\newblock A papier-m{\^a}ch{\'e} approach to learning 3d surface generation.
\newblock In {\em Proceedings of the IEEE conference on computer vision and
  pattern recognition}, pages 216--224, 2018.

\bibitem{hane2017hierarchical}
Christian H{\"a}ne, Shubham Tulsiani, and Jitendra Malik.
\newblock Hierarchical surface prediction for 3d object reconstruction.
\newblock In {\em 2017 International Conference on 3D Vision (3DV)}, pages
  412--420. IEEE, 2017.

\bibitem{hao2020dualsdf}
Zekun Hao, Hadar Averbuch-Elor, Noah Snavely, and Serge Belongie.
\newblock Dualsdf: Semantic shape manipulation using a two-level
  representation.
\newblock In {\em Proceedings of the IEEE/CVF Conference on Computer Vision and
  Pattern Recognition}, pages 7631--7641, 2020.

\bibitem{henderson2018learning}
Paul Henderson and Vittorio Ferrari.
\newblock Learning to generate and reconstruct 3d meshes with only 2d
  supervision.
\newblock {\em arXiv preprint arXiv:1807.09259}, 2018.

\bibitem{kanazawa2018learning}
Angjoo Kanazawa, Shubham Tulsiani, Alexei~A Efros, and Jitendra Malik.
\newblock Learning category-specific mesh reconstruction from image
  collections.
\newblock In {\em Proceedings of the European Conference on Computer Vision
  (ECCV)}, pages 371--386, 2018.

\bibitem{kar2015category}
Abhishek Kar, Shubham Tulsiani, Joao Carreira, and Jitendra Malik.
\newblock Category-specific object reconstruction from a single image.
\newblock In {\em Proceedings of the IEEE conference on computer vision and
  pattern recognition}, pages 1966--1974, 2015.

\bibitem{kato2018neural}
Hiroharu Kato, Yoshitaka Ushiku, and Tatsuya Harada.
\newblock Neural 3d mesh renderer.
\newblock In {\em Proceedings of the IEEE Conference on Computer Vision and
  Pattern Recognition}, pages 3907--3916, 2018.

\bibitem{kazhdan2003rotation}
Michael Kazhdan, Thomas Funkhouser, and Szymon Rusinkiewicz.
\newblock Rotation invariant spherical harmonic representation of 3 d shape
  descriptors.
\newblock In {\em Symposium on geometry processing}, volume~6, pages 156--164,
  2003.

\bibitem{kingma2013auto}
Diederik~P Kingma and Max Welling.
\newblock Auto-encoding variational bayes.
\newblock {\em arXiv preprint arXiv:1312.6114}, 2013.

\bibitem{kleineberg2020adversarial}
Marian Kleineberg, Matthias Fey, and Frank Weichert.
\newblock Adversarial generation of continuous implicit shape representations.
\newblock {\em arXiv preprint arXiv:2002.00349}, 2020.

\bibitem{li2018point}
Chun-Liang Li, Manzil Zaheer, Yang Zhang, Barnabas Poczos, and Ruslan
  Salakhutdinov.
\newblock Point cloud gan.
\newblock {\em arXiv preprint arXiv:1810.05795}, 2018.

\bibitem{li2018differentiable}
Tzu-Mao Li, Miika Aittala, Fr{\'e}do Durand, and Jaakko Lehtinen.
\newblock Differentiable monte carlo ray tracing through edge sampling.
\newblock {\em ACM Transactions on Graphics (TOG)}, 37(6):1--11, 2018.

\bibitem{lim2017geometric}
Jae~Hyun Lim and Jong~Chul Ye.
\newblock Geometric gan.
\newblock {\em arXiv preprint arXiv:1705.02894}, 2017.

\bibitem{liu2019soft}
Shichen Liu, Tianye Li, Weikai Chen, and Hao Li.
\newblock Soft rasterizer: A differentiable renderer for image-based 3d
  reasoning.
\newblock In {\em Proceedings of the IEEE International Conference on Computer
  Vision}, pages 7708--7717, 2019.

\bibitem{liu2020dist}
Shaohui Liu, Yinda Zhang, Songyou Peng, Boxin Shi, Marc Pollefeys, and Zhaopeng
  Cui.
\newblock Dist: Rendering deep implicit signed distance function with
  differentiable sphere tracing.
\newblock In {\em Proceedings of the IEEE/CVF Conference on Computer Vision and
  Pattern Recognition}, pages 2019--2028, 2020.

\bibitem{loper2014opendr}
Matthew~M Loper and Michael~J Black.
\newblock Opendr: An approximate differentiable renderer.
\newblock In {\em European Conference on Computer Vision}, pages 154--169.
  Springer, 2014.

\bibitem{lorensen1987marching}
William~E Lorensen and Harvey~E Cline.
\newblock Marching cubes: A high resolution 3d surface construction algorithm.
\newblock {\em ACM siggraph computer graphics}, 21(4):163--169, 1987.

\bibitem{luo2020end}
Andrew Luo, Zhoutong Zhang, Jiajun Wu, and Joshua~B Tenenbaum.
\newblock End-to-end optimization of scene layout.
\newblock In {\em Proceedings of the IEEE/CVF Conference on Computer Vision and
  Pattern Recognition}, pages 3754--3763, 2020.

\bibitem{maas2013rectifier}
Andrew~L Maas, Awni~Y Hannun, and Andrew~Y Ng.
\newblock Rectifier nonlinearities improve neural network acoustic models.
\newblock In {\em Proc. icml}, volume~30, page~3. Citeseer, 2013.

\bibitem{makhzani2015adversarial}
Alireza Makhzani, Jonathon Shlens, Navdeep Jaitly, Ian Goodfellow, and Brendan
  Frey.
\newblock Adversarial autoencoders.
\newblock {\em arXiv preprint arXiv:1511.05644}, 2015.

\bibitem{mescheder2019occupancy}
Lars Mescheder, Michael Oechsle, Michael Niemeyer, Sebastian Nowozin, and
  Andreas Geiger.
\newblock Occupancy networks: Learning 3d reconstruction in function space.
\newblock In {\em Proceedings of the IEEE Conference on Computer Vision and
  Pattern Recognition}, pages 4460--4470, 2019.

\bibitem{mildenhall2020nerf}
Ben Mildenhall, Pratul~P Srinivasan, Matthew Tancik, Jonathan~T Barron, Ravi
  Ramamoorthi, and Ren Ng.
\newblock Nerf: Representing scenes as neural radiance fields for view
  synthesis.
\newblock In {\em European Conference on Computer Vision}, pages 405--421.
  Springer, 2020.

\bibitem{niemeyer2020differentiable}
Michael Niemeyer, Lars Mescheder, Michael Oechsle, and Andreas Geiger.
\newblock Differentiable volumetric rendering: Learning implicit 3d
  representations without 3d supervision.
\newblock In {\em Proceedings of the IEEE/CVF Conference on Computer Vision and
  Pattern Recognition}, pages 3504--3515, 2020.

\bibitem{park2019deepsdf}
Jeong~Joon Park, Peter Florence, Julian Straub, Richard Newcombe, and Steven
  Lovegrove.
\newblock Deepsdf: Learning continuous signed distance functions for shape
  representation.
\newblock In {\em Proceedings of the IEEE Conference on Computer Vision and
  Pattern Recognition}, pages 165--174, 2019.

\bibitem{samuel_f_potter_2021_4609402}
Samuel~F. Potter.
\newblock sampotter/python-embree:, Mar. 2021.

\bibitem{ravi2020accelerating}
Nikhila Ravi, Jeremy Reizenstein, David Novotny, Taylor Gordon, Wan-Yen Lo,
  Justin Johnson, and Georgia Gkioxari.
\newblock Accelerating 3d deep learning with pytorch3d.
\newblock {\em arXiv preprint arXiv:2007.08501}, 2020.

\bibitem{remelli2020meshsdf}
Edoardo Remelli, Artem Lukoianov, Stephan~R Richter, Beno{\^\i}t Guillard,
  Timur Bagautdinov, Pierre Baque, and Pascal Fua.
\newblock Meshsdf: Differentiable iso-surface extraction.
\newblock {\em arXiv preprint arXiv:2006.03997}, 2020.

\bibitem{salimans2016improved}
Tim Salimans, Ian Goodfellow, Wojciech Zaremba, Vicki Cheung, Alec Radford, and
  Xi Chen.
\newblock Improved techniques for training gans.
\newblock {\em Advances in neural information processing systems},
  29:2234--2242, 2016.

\bibitem{shu20193d}
Dong~Wook Shu, Sung~Woo Park, and Junseok Kwon.
\newblock 3d point cloud generative adversarial network based on tree
  structured graph convolutions.
\newblock In {\em Proceedings of the IEEE International Conference on Computer
  Vision}, pages 3859--3868, 2019.

\bibitem{sitzmann2020metasdf}
Vincent Sitzmann, Eric~R Chan, Richard Tucker, Noah Snavely, and Gordon
  Wetzstein.
\newblock Metasdf: Meta-learning signed distance functions.
\newblock {\em arXiv preprint arXiv:2006.09662}, 2020.

\bibitem{sitzmann2019scene}
Vincent Sitzmann, Michael Zollh{\"o}fer, and Gordon Wetzstein.
\newblock Scene representation networks: Continuous 3d-structure-aware neural
  scene representations.
\newblock {\em arXiv preprint arXiv:1906.01618}, 2019.

\bibitem{smith2019geometrics}
Edward~J Smith, Scott Fujimoto, Adriana Romero, and David Meger.
\newblock Geometrics: Exploiting geometric structure for graph-encoded objects.
\newblock {\em arXiv preprint arXiv:1901.11461}, 2019.

\bibitem{tewari2018self}
Ayush Tewari, Michael Zollh{\"o}fer, Pablo Garrido, Florian Bernard, Hyeongwoo
  Kim, Patrick P{\'e}rez, and Christian Theobalt.
\newblock Self-supervised multi-level face model learning for monocular
  reconstruction at over 250 hz.
\newblock In {\em Proceedings of the IEEE Conference on Computer Vision and
  Pattern Recognition}, pages 2549--2559, 2018.

\bibitem{tran2017deep}
Dustin Tran, Rajesh Ranganath, and David~M Blei.
\newblock Deep and hierarchical implicit models.
\newblock {\em arXiv preprint arXiv:1702.08896}, 7(3):13, 2017.

\bibitem{ulyanov2016instance}
Dmitry Ulyanov, Andrea Vedaldi, and Victor Lempitsky.
\newblock Instance normalization: The missing ingredient for fast stylization.
\newblock {\em arXiv preprint arXiv:1607.08022}, 2016.

\bibitem{valsesia2018learning}
Diego Valsesia, Giulia Fracastoro, and Enrico Magli.
\newblock Learning localized generative models for 3d point clouds via graph
  convolution.
\newblock In {\em International conference on learning representations}, 2018.

\bibitem{vranic2001tools}
Dejan~V Vranic, Dietmar Saupe, and J{\"o}rg Richter.
\newblock Tools for 3d-object retrieval: Karhunen-loeve transform and spherical
  harmonics.
\newblock In {\em 2001 IEEE Fourth Workshop on Multimedia Signal Processing
  (Cat. No. 01TH8564)}, pages 293--298. IEEE, 2001.

\bibitem{wald2014embree}
Ingo Wald, Sven Woop, Carsten Benthin, Gregory~S Johnson, and Manfred Ernst.
\newblock Embree: a kernel framework for efficient cpu ray tracing.
\newblock {\em ACM Transactions on Graphics (TOG)}, 33(4):1--8, 2014.

\bibitem{wang2019mvpnet}
Jinglu Wang, Bo Sun, and Yan Lu.
\newblock Mvpnet: Multi-view point regression networks for 3d object
  reconstruction from a single image.
\newblock In {\em Proceedings of the AAAI Conference on Artificial
  Intelligence}, volume~33, pages 8949--8956, 2019.

\bibitem{wang2018pixel2mesh}
Nanyang Wang, Yinda Zhang, Zhuwen Li, Yanwei Fu, Wei Liu, and Yu-Gang Jiang.
\newblock Pixel2mesh: Generating 3d mesh models from single rgb images.
\newblock In {\em Proceedings of the European Conference on Computer Vision
  (ECCV)}, pages 52--67, 2018.

\bibitem{wenninger1999spherical}
Magnus~J Wenninger.
\newblock {\em Spherical models}, volume~3.
\newblock Courier Corporation, 1999.

\bibitem{williams2019deep}
Francis Williams, Teseo Schneider, Claudio Silva, Denis Zorin, Joan Bruna, and
  Daniele Panozzo.
\newblock Deep geometric prior for surface reconstruction.
\newblock In {\em Proceedings of the IEEE Conference on Computer Vision and
  Pattern Recognition}, pages 10130--10139, 2019.

\bibitem{wu2017marrnet}
Jiajun Wu, Yifan Wang, Tianfan Xue, Xingyuan Sun, Bill Freeman, and Josh
  Tenenbaum.
\newblock Marrnet: 3d shape reconstruction via 2.5 d sketches.
\newblock In {\em Advances in neural information processing systems}, pages
  540--550, 2017.

\bibitem{wu2016learning}
Jiajun Wu, Chengkai Zhang, Tianfan Xue, Bill Freeman, and Josh Tenenbaum.
\newblock Learning a probabilistic latent space of object shapes via 3d
  generative-adversarial modeling.
\newblock {\em Advances in neural information processing systems}, 29:82--90,
  2016.

\bibitem{wu20153d}
Zhirong Wu, Shuran Song, Aditya Khosla, Fisher Yu, Linguang Zhang, Xiaoou Tang,
  and Jianxiong Xiao.
\newblock 3d shapenets: A deep representation for volumetric shapes.
\newblock In {\em Proceedings of the IEEE conference on computer vision and
  pattern recognition}, pages 1912--1920, 2015.

\bibitem{xu2019disn}
Qiangeng Xu, Weiyue Wang, Duygu Ceylan, Radomir Mech, and Ulrich Neumann.
\newblock Disn: Deep implicit surface network for high-quality single-view 3d
  reconstruction.
\newblock In {\em Advances in Neural Information Processing Systems}, pages
  492--502, 2019.

\bibitem{yan2016perspective}
Xinchen Yan, Jimei Yang, Ersin Yumer, Yijie Guo, and Honglak Lee.
\newblock Perspective transformer nets: Learning single-view 3d object
  reconstruction without 3d supervision.
\newblock In {\em Advances in neural information processing systems}, pages
  1696--1704, 2016.

\bibitem{mcubes_pytorch}
Tatsuya Yatagawa.
\newblock mcubes\_pytorch: Pytorch implementation for marching cubes.

\bibitem{yu20033d}
Meng Yu, Indriyati Atmosukarto, Wee~Kheng Leow, Zhiyong Huang, and Rong Xu.
\newblock 3d model retrieval with morphing-based geometric and topological
  feature maps.
\newblock In {\em 2003 IEEE Computer Society Conference on Computer Vision and
  Pattern Recognition, 2003. Proceedings.}, volume~2, pages II--656. IEEE,
  2003.

\bibitem{zadeh2019variational}
Amir Zadeh, Yao-Chong Lim, Paul~Pu Liang, and Louis-Philippe Morency.
\newblock Variational auto-decoder.
\newblock {\em arXiv preprint arXiv:1903.00840}, 2019.

\bibitem{zamorski2018adversarial}
Maciej Zamorski, Maciej Zieba, Piotr Klukowski, Rafa{\l} Nowak, Karol Kurach,
  Wojciech Stokowiec, and Tomasz Trzci{\'n}ski.
\newblock Adversarial autoencoders for compact representations of 3d point
  clouds.
\newblock {\em arXiv preprint arXiv:1811.07605}, 2018.

\bibitem{zhang2018learning}
Xiuming Zhang, Zhoutong Zhang, Chengkai Zhang, Josh Tenenbaum, Bill Freeman,
  and Jiajun Wu.
\newblock Learning to reconstruct shapes from unseen classes.
\newblock {\em Advances in neural information processing systems},
  31:2257--2268, 2018.

\end{thebibliography}
}
\newpage
\setcounter{section}{0}
\setcounter{table}{0}
\setcounter{figure}{0}
\onecolumn
\begin{center}\textbf{\Large{Supplementary material}}\end{center}
\section{Training of our SurfGen Model}

\noindent We implement out SurfGen model following the steps detailed in Algorithms 1, 2, and 3. Where $f_{\mathcal{M} \rightarrow S}$ is a differentiable function which computes the spherical surface features from an input mesh.\\
\begin{multicols}{2}{
\begin{algorithm}[H]
\SetAlgoLined
\textbf{Input:} Mean \& std for shape $I$: $\mu_I, \sigma_I$\\
\hspace{10.7mm}Points set sampled for shape $I$: $P_I$ \\
\hspace{10.7mm}Ground truth signed distance for $P_I$: $S_{GT}$ \\

1: Sample latent code $z \sim \mathcal{N}(\mu_{I}, \sigma^2_I\cdot\mathbb{I})$\\
2: Assemble 3D sampling grid $G_{3D}$\\
3: Sample field for $G_{3D}$ as $S_G = g_{\psi}(G_{3D}, z)$\\
4: Extract isosurface $\mathcal{M}=(V,F) = \text{MC}(S_{G}, G_{3D})$\\
5: Spherical projection $\text{Sph}_{\mathcal{M}} = f_{\mathcal{M} \rightarrow S}(\mathcal{M})$\\
6: Sample field for $P_I$ as $S_P = g_{\psi}(P_I, z)$\\
7: Adversarial loss: $\mathcal{L}_{\text{GAN}} = -D(Sph_{\mathcal{M}})$\\
8: Clamped L1 loss: $\mathcal{L}_{\text{SDF}} = {L_{1,{\pm T}}(S_{GT}, S_P)}$\\
9: KLD regularization loss: $D_{KL}(q(z|x)||p(z))$\\
\textbf{Output:} Total loss $\mathcal{L} = \alpha \mathcal{L}_{\text{GAN}} + \beta \mathcal{L}_{\text{SDF}} + \gamma D_{\text{KL}}$ 

\caption{Forward pass of generator}
\end{algorithm}
\begin{algorithm}[H]
\SetAlgoLined
\textbf{Input:} Upstream gradient on spherical $\frac{\partial \mathcal{L}_{\text{GAN}}}{\partial Sph_{\mathcal{M}}}$ \\
\hspace{10.7mm}Clamped L1 loss: $\mathcal{L}_{\text{SDF}}$\\
\hspace{10.7mm}KLD regularization loss $D_{\text{KL}}$ \\
\hspace{10.7mm}Latent code $z$\\
1: Compute GAN gradient w.r.t. $V$: $\frac{\partial \mathcal{L}_{\text{GAN}}}{\partial Sph_{\mathcal{M}}} \frac{\partial Sph_{\mathcal{M}}}{\partial V}$\\
2: Forward pass $S_v = g_{\psi}(v, z)$ for $v \in V$ \\
3: Find normal $n(v) = \nabla g_{\psi}(v, z)$ for $v \in V$\\
4: $\frac{\partial \mathcal{L}_{\text{GAN}}}{\partial g_{\psi}}(v) = -\frac{\partial \mathcal{L}_{\text{GAN}}}{\partial v}n(v)$ for $v \in V$\\
5: $\frac{\partial \mathcal{L}_{\text{GAN}}}{\partial z} = \sum_{v \in V} \frac{\partial \mathcal{L}_{\text{GAN}}}{\partial g_{\psi}}(v) \frac{ \partial g_{\psi}}{\partial z}(v)$\\
6: Gradient from L1: $\frac{\partial \mathcal{L}_{\text{SDF}}}{\partial g_{\psi}}$, $\frac{\partial \mathcal{L}_{\text{SDF}}}{\partial \mu}$, $\frac{\partial \mathcal{L}_{\text{SDF}}}{\partial \sigma}$\\
7: Gradient from KLD: $\frac{\partial D_{\text{KL}}}{\partial \mu}$, $\frac{\partial D_{\text{KL}}}{\partial \sigma}$\\

8: $\psi  \leftarrow \psi - \alpha \frac{\partial \mathcal{L}_{\text{GAN}}}{\partial g_{\psi}} - \beta \frac{\partial \mathcal{L}_{\text{SDF}}}{\partial g_{\psi}}$ \\
9: $\mu \leftarrow \mu - \alpha \frac{\partial \mathcal{L}_{\text{GAN}}}{\partial \mu} - \beta \frac{\partial \mathcal{L}_{\text{SDF}}}{\partial \mu} - \gamma \frac{\partial D_{\text{KL}}}{\partial \mu}$ \\
10:$\sigma \leftarrow \sigma - \alpha \frac{\partial \mathcal{L}_{\text{GAN}}}{\partial \sigma} - \beta \frac{\partial \mathcal{L}_{\text{SDF}}}{\partial \sigma} - \gamma \frac{\partial D_{\text{KL}}}{\partial \sigma}$  \\
\textbf{Output: } Updated generator \& embedding\\
\caption{Backwards pass of generator}
\end{algorithm}
}
{\columnbreak

\begin{algorithm}[H]
\SetAlgoLined
\textbf{Input:} Spherical projection of real shape $\text{Sph}_{\text{real}}$,\\
\hspace{10.7mm}Spherical projection of fake shape $\text{Sph}_{\mathcal{M}}$,\\
\hspace{10.7mm}Discriminator $\text{D}_{\phi}$\\
1: $\mathcal{L}_{\text{D}} = \text{ReLU}(1-\text{D}_{\phi}(\text{Sph}_{\text{real}}))$\\ $\hspace{9.4mm}+ \hspace{1mm} \text{ReLU}(1 + \text{D}_{\phi}(\text{Sph}_{\mathcal{M}}))$\\
2: $\phi \leftarrow \phi - \frac{\partial }{\partial \phi}\mathcal{L}_{\text{D}}$\\
\textbf{Output:} Updated discriminator $\text{D}_{\phi}$

\caption{Training of discriminator}
\end{algorithm}

}
\end{multicols}

\newpage
\section{Derivation of depth projection}
\noindent If a ray $i$ directly hits inside a triangle $j$ at $p_{j^{i}}$, with euclidean coordinates of $\overrightarrow{p_{j^{i}}} = (x, y, z)$, we can express $\overrightarrow{p_{j^{i}}}$ differentiably as a weighted sum of triangle vertices $\overrightarrow{v_{j,0}}, \overrightarrow{v_{j,1}},\overrightarrow{v_{j,2}}$ in barycentric coordinates: 
\begin{align}
\overrightarrow{p_{j^{i}}} &= w^i_{j,0} \cdot \overrightarrow{v_{j,0}} + w^i_{j,1} \cdot \overrightarrow{v_{j,1}} + w^i_{j,2} \cdot \overrightarrow{v_{j,2}}~\label{one}\\
 1&=  w^i_{j,0} + w^i_{j,1} + w^i_{j,2}\label{two}
\end{align}
\noindent $(w^i_{j,0}, w^i_{j,1}, w^i_{j,2})$ can be expressed in a way that is differentiable w.r.t. triangle vertices using the ratio of the area between the three subtriangles created:
\begin{align}
\text{Area}_j &= 0.5 * ||  (\overrightarrow{v_{j,1}}-\overrightarrow{v_{j,0}}) \times (\overrightarrow{v_{j,2}}-\overrightarrow{v_{j,0}})||_2\\
w^i_{j,0}&= 0.5* ||(\overrightarrow{v_{j,2}}-\overrightarrow{v_{j,1}}) \times (\overrightarrow{p_{j^{i}}} - \overrightarrow{v_{j,1}})||_2 / \text{Area}_j\\
w^i_{j,1}&= 0.5* ||(\overrightarrow{v_{j,0}}-\overrightarrow{v_{j,2}}) \times (\overrightarrow{p_{j^{i}}} - \overrightarrow{v_{j,2}})||_2 / \text{Area}_j\\
w^i_{j,2}&= 1-w^i_{j,1}-w^i_{j,0}
\end{align}
\noindent where $\times$ denotes cross product, and we are assuming the ordering of the vertices of a triangle follows the right-hand rule.

\section{Derivation of near miss distance}
\noindent If a ray does not intersect with the inside of a triangle, but  is within $r$ distance of a triangle, we need to compute the distance between near miss point $\overrightarrow{p_{j^{i}}}$ on the ray and the closest point on the triangle $F_j$. Assume the ray has a closest point on the triangle edge defined by $\overrightarrow{v_{j,1}}, \overrightarrow{v_{j,0}}$:
\begin{align}
n &= ||\overrightarrow{v_{j,1}}-\overrightarrow{v_{j,0}}||^2_2\\
t &= \text{clamp}((\overrightarrow{p_{j^{i}}} - \overrightarrow{v_{j,0}})\cdot (\overrightarrow{v_{j,1}}-\overrightarrow{v_{j,0}})/n, \text{min}=0.0, \text{max}=1.0)\\
\text{nn}_\text{point} &= \overrightarrow{v_{j,0}} + t * (\overrightarrow{v_{j,1}}-\overrightarrow{v_{j,0}})\\
\text{nn}_\text{dist} &= ||\overrightarrow{p_{j^{i}}}-\text{nn}_\text{point}||_2
\end{align}
where $\cdot$ denotes dot product. Note that in practice we use squared euclidean distance $||\cdot||^2_2$.

\newpage
\section{Visualization of learned discriminator features}
\begin{figure*}[h!]
    \centering
    \includegraphics[width=1.0\linewidth]{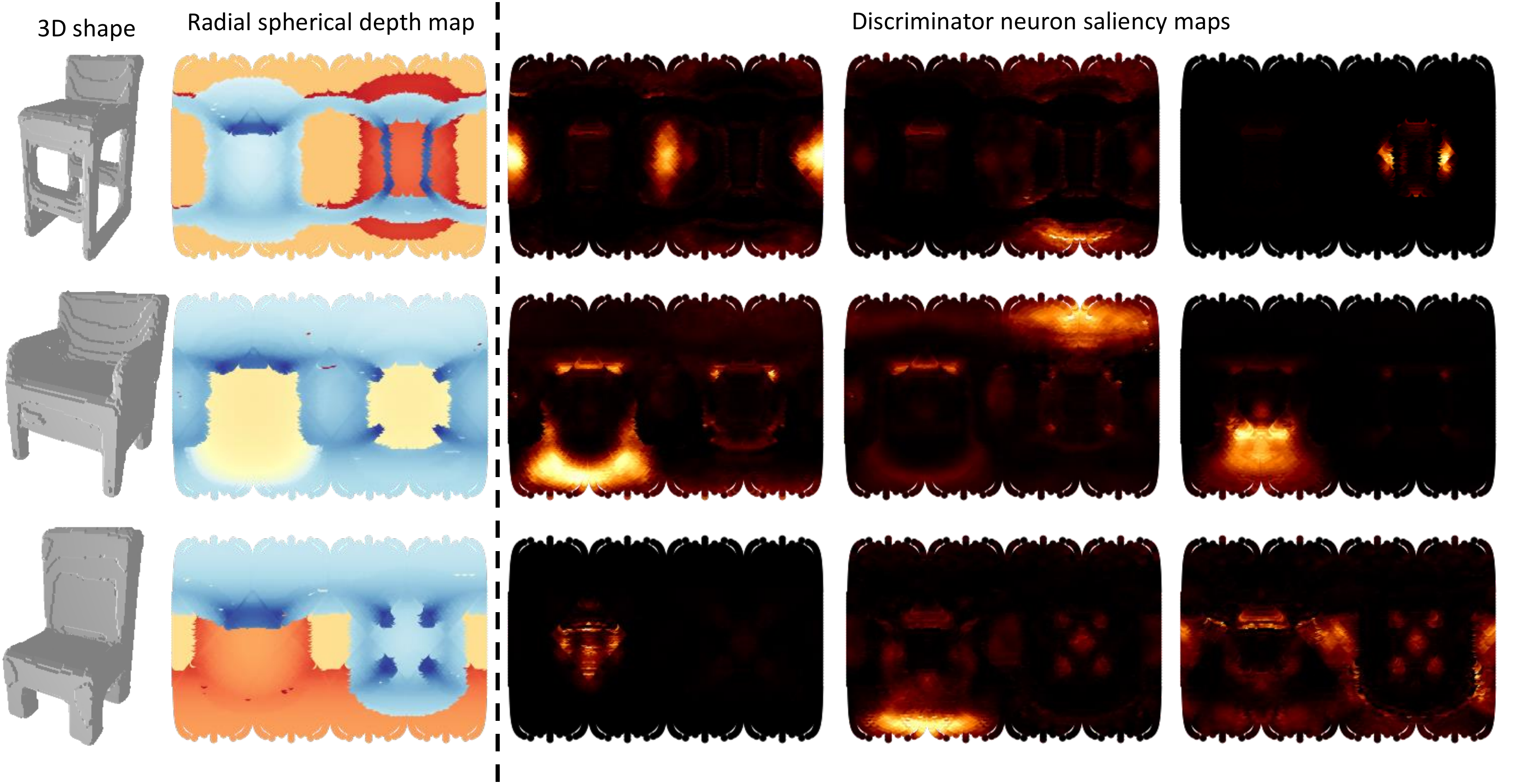}
    \caption{We visualize the individual salient features from the last 256 neuron fully connected layer of the discriminator. Gradients are visualized using guided backpropagation. From left to right: 3D shape, radial spherical depth map flattened, and salient features for three neurons for each shape. Note that not the same neurons are show for different shapes.}
    \label{saliency}
\end{figure*}

We visualize the features learned by our discriminator using guided backpropagation in Figure~\ref{saliency}. It can be seen that individual neurons can learn localized features that are specific to a given geometry in each shape.

\end{document}